\documentclass[10pt,twocolumn,letterpaper]{article}

\usepackage{cvpr}              

\usepackage{graphicx}
\usepackage{amsmath}
\usepackage{amssymb}
\usepackage{booktabs}
\usepackage{multirow}
\usepackage{colortbl}
\usepackage{makecell}

\usepackage[dvipsnames]{xcolor}

\definecolor{cvprblue}{rgb}{0.21,0.49,0.74}
\usepackage[pagebackref,breaklinks,colorlinks,citecolor=cvprblue]{hyperref}

\usepackage[capitalize]{cleveref}
\crefname{section}{Sec.}{Secs.}
\Crefname{section}{Section}{Sections}
\Crefname{table}{Table}{Tables}
\crefname{table}{Tab.}{Tabs.}

\def\cvprPaperID{4448} 
\def\confName{CVPR}
\def\confYear{2024}

\begin{document}

\title{OpticalDR: A Deep Optical Imaging Model for Privacy-Protective \\Depression Recognition}

\author{
    {Yuchen Pan, Junjun Jiang\thanks{Correspondence to: Junjun Jiang (junjunjiang@hit.edu.cn)} , Kui Jiang, Zhihao Wu, Keyuan Yu, Xianming Liu}\\
    {Harbin Institute of Technology}\\
}

\maketitle

\begin{abstract}
    Depression Recognition (DR) poses a considerable challenge, especially in the context of the growing concerns surrounding privacy. Traditional automatic diagnosis of DR technology necessitates the use of facial images, undoubtedly expose the patient identity features and poses privacy risks. In order to mitigate the potential risks associated with the inappropriate disclosure of patient facial images, we design a new imaging system to erase the identity information of captured facial images while retain disease-relevant features. It is irreversible for identity information recovery while preserving essential disease-related characteristics necessary for accurate DR. 
    More specifically, we try to record a de-identified facial image (erasing the identifiable features as much as possible) by a learnable lens, which is optimized in conjunction with the following DR task as well as a range of face analysis related auxiliary tasks in an end-to-end manner. 
    These aforementioned strategies form our final Optical deep Depression Recognition network (OpticalDR). Experiments on CelebA, AVEC 2013, and AVEC 2014 datasets demonstrate that our OpticalDR has achieved state-of-the-art privacy protection performance with an average AUC of 0.51 on popular facial recognition models, and competitive results for DR with MAE/RMSE of 7.53/8.48 on AVEC 2013 and 7.89/8.82 on AVEC 2014, respectively. Code is available at https://github.com/divertingPan/OpticalDR.
\end{abstract}

\section{Introduction}
\label{sec:intro}

\begin{figure}[t]
\centering
		\centerline{\includegraphics[width=\linewidth]{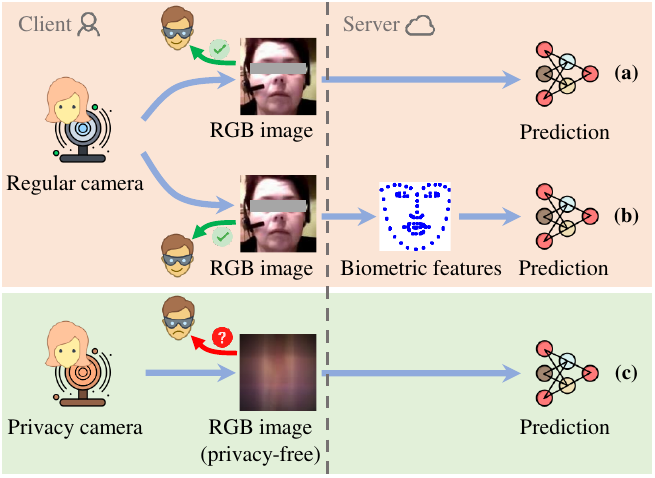}}
		\caption{Different DR approaches: (a) conventional facial recognition with no privacy preservation, (b) facial features-based approaches with limited privacy preservation, and (c) our full privacy-preserving approach that doesn't generate facial images. The DR approaches with regular cameras could be at the risk of sensitive information access by attackers after the images are captured. However, with our approach, no sensitive information would be captured or stored, whether on the client or the server. }
		\label{fig_0}
\end{figure}

Automated depression diagnosis (ADD) approaches have emerged based on various biometric features, including acoustics, expressions, and electroencephalogram \cite{2.4, DONG2021279, 4.13, 10097883}.
Among these approaches, facial-based ADD has become mainstream \cite{2018arXiv181108592H, 8756584, 10185131}, offering the advantages of easy access and non-contact assessment. However, facial images from depression patients are particularly sensitive, as they inherently encompass biometric identifiers and pose privacy risk.
Consequently, it is of paramount importance to address the safeguarding of facial information when developing ADD systems to uphold medical privacy and security.


To minimize these risks, previous studies \cite{2018arXiv181108592H, 8756584, 10185131} 
introduce facial image anonymization techniques. Rather than the full facial images, facial biometric features (identity-protected) such as facial landmarks are used for DR (Fig.~\ref{fig_0}). 
These endeavors, however, are still dependent on the presence of visually accessible images for feature extraction, which poses a significant threat to privacy. Therefore, it is crucial to develop a more secure and robust facial imaging approach at the initial stage of data collection. 

In the field of Deep Optics \cite{optics01}, 
the lens can be parameterized to a distinct layer within the deep learning model. Inspired by this, in this paper we integrate the Deep Optics into deep learning network and advises an Optical deep DR network – which we term a OpticalDR – that provides a graceful solution for privacy-protective DR.
Compared to the fixed blurring or defocus image processing methods, which might compromise depression-related features when removing privacy information, the proposed lens is learnable and can be optimized in an end-to-end manner, like a specialized feature extractor on the front of deep network. 
As demonstrated in Fig.~\ref{fig_1}, we develop a learnable single-lens optical model to extract depression-related features while erasing sensitive privacy information from the observation. 
In the subsequent deep learning model, we design a progressive learning tactic to train our network. To harness abundant supervision (label) information from various face datasets, we incrementally incorporate different tasks, \emph{i.e.},identity recognition, emotion recognition and DR, to assist our deep DR network training. It is worth noting that acquiring a complete dataset with labels for identity, emotion, and depression is extremely challenging in practice. Consequently, the adoption of progressive learning offers a viable strategy to leverage the available supervision information from existing datasets. Our contributions can be summarized as follows:

\begin{enumerate}
	
    \item We have successfully simulated a privacy lens with parameterized model. The simulated lens is co-optimized alongside the subsequent model to ensure facial privacy protection while retaining valuable features.
	
    \item We introduce OpticalDR, a novel framework that integrates an optical lens with a deep learning model to extract depression-related information from privacy-preserving images. The entire pipeline operates without the need for storing or transmitting facial images, ensuring a heightened level of privacy protection.
	
    \item Our experimental results demonstrate competitiveness on the AVEC 2013 and AVEC 2014 datasets in comparison to existing DR approaches. Furthermore, the optical component in OpticalDR exhibits robustness against various types of attacks.
	
\end{enumerate}

\section{Related Works}

\paragraph{Visual Depression Recognition.} Contemporary research in DR primarily relies on audio and video signals obtained from volunteers during interviews. Datasets such as AVEC 2013 \cite{avec13} and AVEC 2014 \cite{avec14} contain original facial images within video recordings. Some studies \cite{PAN2024121410, 9186792} have concentrated on recognizing depression from video frames that contain the original facial images. Early approaches were centered on hand-crafted designed feature extraction and appropriate regression models \cite{avec13, 8501575} for DR using visual data. Growing concerns about privacy have emerged. As a result, the AVEC 2017 \cite{avec17} dataset takes a privacy-centric approach, providing only facial landmarks and other manually engineered facial features. Some research employed facial landmarks and other physiological data as facial features, often considering the motion of landmark points \cite{E17V2} or incorporating factors like Action Units and head pose \cite{8756584, E17V1}. Furthermore, studies have shown that human behavior representations \cite{8976305, 8373825} can significantly contribute to DR from facial data. More recently, deep learning-based methods \cite{8344107, 9187982}, or a combination of hand-crafted and deep features \cite{8976084, 9667301}, have been applied, resulting in substantial improvements in recognition performance. Nevertheless, approaches relying on facial features may not necessitate the use of original images during inference, but these features still need to be extracted from facial images\textbf{}, exposing potential risks of facial data leakage.

\paragraph{Deep Optics for Privacy Protection.} 
Recently, Deep Optics \cite{optics01} has gained more of the spotlight in research communities with the advancement of deep learning. It involves the joint optimization of optical systems and downstream tasks, and has demonstrated success in various computer vision tasks such as depth estimation \cite{9466261}, 3D object detection \cite{9010976}, extended depth of field, super-resolution imaging \cite{optics01}, and image classification \cite{Chang2018}. The underlying philosophy is to capture the downstream task friendly images to improve the performance of high-level applications.


To achieve a privacy protection high-level visual task, the optical imaging system is designed to intentionally degrade image quality and obscure private information, while still enabling downstream tasks. This approach has been successfully demonstrated in lensless camera facial recognition \cite{optics04, 10096627}, deep optimized lens cameras for action recognition \cite{optics02}, and pose estimation \cite{9710270}. This work focuses on exploring facial privacy protection using Deep Optics, primarily on facial images. The goal is to extract facial depression features from privacy-preserving images for a DR system with enhanced privacy.


\section{The Proposed Method}

\begin{figure*}[t]
\centering
		\centerline{\includegraphics[width=\linewidth]{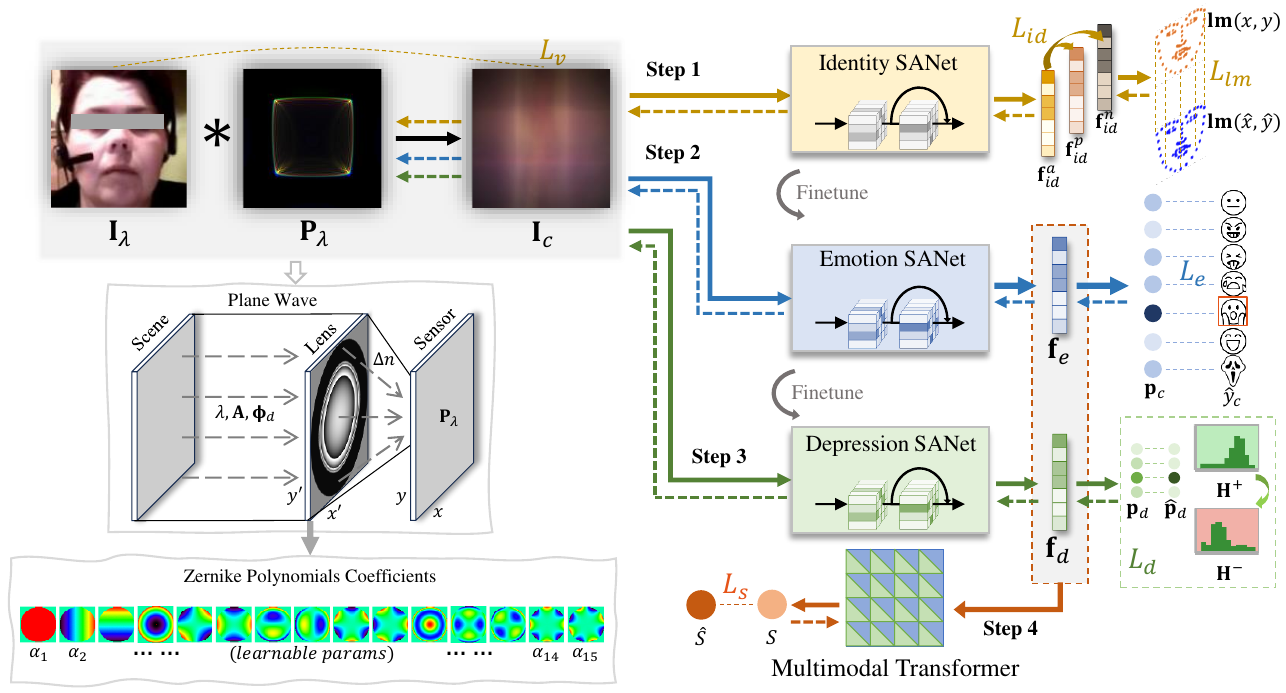}}
		\caption{The architecture of the proposed OpticalDR. The optical model comprises a thin lens for image capture in front of the sensor. The deep learning model utilizes the captured image and is jointly optimized with the optical component for depression score recognition.}
		\label{fig_1}
\end{figure*}

We are primarily focused on the task of privacy-preserving DR. The main objective is to create a personal information erased ``face'' image using a specially optimized lens, which is achieved through joint optimization with the DR model. Therefore, OpticalDR offers privacy preservation at a hardware level, ensuring that no privacy information is generated or stored in digital devices. This entire process is illustrated in Fig.~\ref{fig_1}. The OpticalDR comprises two key components: the optical model and the deep feature extraction model. In the optical model, we parameterize a learnable single optical element for generating privacy preserving image (Sec. \ref{sec:OpticalModel}). The deep feature model is responsible for the subsequent extraction of emotional and depression features and the fusion of these features for the final prediction (Sec. \ref{sec:Privacy-free}). Following that, we explain the auxiliary tasks and the corresponding loss functions designed for optimizing the OpticalDR (Sec. \ref{sec:loss}). 


\subsection{Privacy Erasing Optical Model}
\label{sec:OpticalModel}
The optical model corresponds to the image generation process captured by the camera sensor. The objective of this model is to produce an image that is visually unrecognizable and doesn't reveal the identity of the subject. Simultaneously, this process must preserve the important features required for DR. This is achieved through end-to-end joint optimization with the feature extraction model, utilizing appropriate loss functions.

To accomplish this, we derive a wave-based image formation model using a single thin lens, as discussed in \cite{optics01}. The key variables in the optical model include the surface profile denoted as $\mathbf{H}$, represented by Zernike term coefficients $\alpha$, and the PSF represented as $\mathbf{P}$ of the lens, which can be derived from $\mathbf{H}$. The image passing through the lens can be calculated from $\mathbf{P}$ using the following equation:

\begin{equation}
	\mathbf{I}_{c}\left(x,y\right)=\int\left(\mathbf{I}_{\lambda}\ast \mathbf{P}_{\lambda}\right)\left(x,y\right)\kappa_{c}\left(\lambda\right)d\lambda + \eta,
\end{equation}
where $\mathbf{I}_{\lambda}$ represents the original scene from the real world with the particular light wavelength $\lambda$ which corresponds to the red, green, and blue light in the RGB image model. $\kappa_{c}$ represents the sensitivity of sensors to these three wavelengths. $\mathbf{I}_{c}$ represents the generated image on the sensor plane. Additionally, we account for sensor noise by introducing the term $\eta$, which is modeled as Gaussian noise with $\eta \sim \mathcal{N}(0,\sigma^{2})$.

The phase change induced by the thin lens is given by

\begin{equation}
	\mathbf{\Phi}\left(x^{\prime},y^{\prime}\right)=\frac{2\pi\Delta n}{\lambda}\mathbf{H}\left(x^{\prime},y^{\prime}\right).
\end{equation}
Here, $\Delta n$ represents the refractive index difference between air and the lens material. $\mathbf{H}$ can be expressed as:

\begin{equation}
	\mathbf{H} = \sum_{j=1}^{q} \alpha_j\mathrm {\mathbf{Z}}_j,
 \label{eq:h}
\end{equation}
where ${\mathbf{Z}}_j$ represents the $j$-th term of the Zernike polynomial in Noll notation \cite{optics05}.





Assuming a point light source at optical infinity with amplitude $\mathbf{A}$ and phase $\mathbf{\Phi}_d$. The PSF $\mathbf{P}$ for the sensor at a distance $z$ can be formulated as follows: 

\begin{equation}
    \mathbf{P}_{\lambda}\left(x,y\right) = \left|\mathcal{F}\left\{\mathbf{A}\left(x',y'\right)e^{i\mathbf{\Phi}\left(x',y'\right)}e^{i\frac{\pi}{\lambda z}\left(x'^{2}+y'^{2}\right)}\right\}\right|^{2}.
\end{equation}

In this way, the PSF $\mathbf{P}$ of the learnable optical element can be simulated through the introduced wave-based image formation module. Since the $\mathbf{P}$ is determined by the optical height map $\mathbf{H}$, we learn from Eq.~\ref{eq:h} that we only need to learn the optimization variable $\{\alpha_j\}_{j=1}^q$.

\subsection{Recognition from Privacy-Preserving Images}
\label{sec:Privacy-free}
In this section, we elaborate on the proposed multi-task deep network for the extraction of intrinsic facial features. In the realm of DR, the incorporation of spatial feature relationships of face image is very important and has shown great efficacy \cite{PAN2024121410, 9667301}. Therefore, we propose to leverage the Spatial Attention Network (SANet) for the extraction of intrinsic facial features. 
Importantly, this module can be seamlessly integrated into any convolutional network architecture. In the context of our SANet implementation, we have chosen to adopt a ResNet \cite{resnet} backbone enhanced with the SA module.

Our approach unfolds in four progressive learning stages. Initially, we pretrain the SANet for recognizing facial features from optically blurred images. This step will harness the benefits of large-scale face recognition database to erase the identity information as well as preserve the facial structure information by the landmark prediction loss. Subsequently, the SANet undergoes fine-tuning to extract emotional features from the blurred image to further extract the DR-related information. Finally, we progressively train the SANet by the DR score loss to represent the depression-related features. The emotional and depression information extracted from these two SANets is harmoniously fused using a multimodal transformer \cite{unalign_trans} layer to obtain the final depression score. 


The fusion of the emotional and depression feature is executed through the multimodal transformer. This operation can be succinctly expressed in the following simplified formulation for the two modalities, representing emotional and depression conditions:

\begin{equation}
	\mathbf{Z}_{\{E,D\}}=[\mathbf{Z}_{E\rightarrow D};\mathbf{Z}_{D\rightarrow E}],
\end{equation}
where $\mathbf{Z}_{E\rightarrow D}$ and $\mathbf{Z}_{D\rightarrow E}$ correspond to the cross-modality attention information, respectively. In this way, the adoption of progressive learning offers an efficient strategy to leverage the available supervision information from different tasks.


\subsection{Joint Optimization of Optical Model and DR} 
\label{sec:loss}

In practical, it is very hard to obtain a full dataset with the labels of identity, emotion and depression, simultaneously. Therefore, in this paper we introduce the progressive learning to train our OpticalDR network, which requires progressively training of the optical lens, a SANet for emotion recognition and a SANet for DR. The final score is predicted by a fusion model with extra training with only the DR dataset. In the following, we will present the details to obtain our DR network. 

\textbf{Step 1: Pretrain the privacy preserving lens.} To achieve privacy preservation, we try to render the images that are visually indistinct from the input image, and unidentifiable by deep learning face recognition models. For visual degradation, we maximize the Mean Squared Error (MSE) between the generated image and the real image, $L_{v}=||\mathbf{I}_c-\mathbf{I}_\lambda||_2^{2}$.

To further obfuscate the identity features in the images, we adopt the inverse triplet loss defined as:

\begin{equation}
	L_{id}=max(d(\mathbf{f}_{id}^a, \mathbf{f}_{id}^n)-d(\mathbf{f}_{id}^a, \mathbf{f}_{id}^p) + m,0),
\end{equation}
where $m$ is a margin value, $\mathbf{f}_{id}^a, \mathbf{f}_{id}^p, \mathbf{f}_{id}^n$ denote the feature embeddings extracted by SANet from the anchor image, the image with the same identity as the anchor, and the image with a different identity than the anchor after passing through the lens. $d(\cdot)$ represents the feature distance. We employ hypersphere distance \cite{9054420} with a margin $r$ to measure it, \emph{i.e.}, $d(x,y)=| \|x\|_{2}^{2} - \|y\|_{2}^{2} | + r$.

Landmarks have proven to be crucial for conveying essential features in DR \cite{10185131}. To ensure that the blurred image retains these landmarks and the vital depression-related features, we introduce a landmark recognition loss $L_{lm}$, which can be calculated by the differences of coordinates of the ground truth and predicted landmarks.  We obtain the landmarks followed the extracted facial embedding and using non-linear mapping from embedding to landmarks.


The total loss for optimizing the identity information can be formulated as:

\begin{equation}
	L_{i}= \alpha L_{v} + \beta L_{id} + \gamma L_{lm},
\end{equation}
where $\alpha, \beta$, and $\gamma$ are the balancing parameters and are set to $-0.1, 1$ and $1$, respectively.

\textbf{Step 2: Integrate emotional information.} In this step, we incorporate emotional information for two primary purposes. First, it serves as emotional prior knowledge to aid in DR, as depression patients often exhibit abnormal emotional arousal. Second, it facilitates the model in learning emotional features from the image captured by the lens. The SANet utilizes the weights obtained from the SANet trained for identity obfuscation in \emph{Step 1} and then is trained to discriminate the facial emotion classes $\hat{y}_c$ with its prediction probability of each class $p_c$, using the following cross-entropy loss:

\begin{equation}
	L_e = -\sum_{c}\hat{y}_c \log(p_{c}).
\end{equation}
The last feature layer of the SANet is retained as the emotional feature $\mathbf{f}_e$.

\textbf{Step 3: Integrate depression information.} To introduce a coarse-grained feature of depression, and bridge the features of emotion and depression, we employ a SANet that utilizes the weights obtained from the SANet trained in \emph{Step 2}. The SANet is fine-tuned to gather information about depression levels. 

In contrast to prior label distribution designs \cite{9187982}, we convert the real-valued depression score into a discrete probability distribution, represented as $\mathbf{p} = [p_1, p_2, \ldots, p_c]$, where $c$ represents the classes of depression levels based on depression scores, as outlined in Tab.~\ref{table_1}. The fine-tuning process minimizes the distribution similarity between the SANet's prediction $\mathbf{p}_d$ and $\mathbf{\hat{p}}_d$, and employs a histogram loss for the label-aware similarity between positive feature pairs (where both samples are from the same depression level) and negative pairs, then define the histogram $\mathbf{H}^+$ and $\mathbf{H}^-$ as defined in \cite{histloss}. Then the depression metric loss can be defined as:

\begin{equation}
	L_d = LAH(\mathbf{\hat{p}}_d, \mathbf{p}_d, \mathbf{H}^+, \mathbf{H}^-).
\end{equation}
The depression feature is extracted from the last feature layer of the SANet and is represented as $\mathbf{f}_d$.

\begin{table}[t]
\centering
\caption{Mapping of depression levels to scores based on the BDI-II criteria in the depression stage training for $L_d$}
\label{table_1}
\resizebox{0.95\linewidth}{!}{
    \begin{tabular}{ccccc}
        \hline
        Score & 0-13        & 14-19    & 20-28        & 29-63      \\ \hline
        Level & 0 (minimal) & 1 (mild) & 2 (moderate) & 3 (severe) \\ \hline
    \end{tabular}
    }
\end{table}

\textbf{Step 4: Fuse emotional and depression features.} Finally, a fusion model is expected to fuse the emotional feature and the coarse-grained depression feature for fine-grained depression score prediction. The complete OpticalDR comprises the pretrained weights of the optics model, the emotion SANet, and the depression SANet. At this stage, the network is frozen, and a final fusion layer is introduced. This layer is trained using the depression score $\hat{S}$ to enable the transformer to learn the mapping from emotional and depression features to the predicted depression score $S$. The optimization of the fusion layer is achieved using the MSE loss:

\begin{equation}
	L_{s}=\sum{\frac{1}{n}}({\hat{S}}-S)^{2}.
\end{equation}

Above-mentioned four steps are sequentially applied to optimize the corresponding part of OpticalDR, as depicted in Fig.~\ref{fig_1}, by the specified loss functions. The emotion SANet is fine-tuned from the identity SANet, and the depression SANet is fine-tuned from the emotion SANet to incorporate prior information. Additional details and experimental settings are provided in the supplementary materials.


\section{Experiments}

\subsection{Datasets}

In our experiments, we utilized the following datasets:
i) \textbf{CelebA} \cite{celeba}: This dataset comprises human face images with corresponding identity labels, totaling 10,177 identities and 202,599 samples. We assigned the last 1,000 samples for the validation set and the remainder for training.
ii) \textbf{CK+} \cite{5543262}: It comprises 593 series of facial images from 123 identities, spanning 7 emotion classes for all samples. We allocated 300 samples for validation and utilized the remainder for training.
iii) \textbf{AVEC 2013} \cite{avec13}: This dataset includes 150 video clips with 82 unique participants, exclusively employed in the validation and testing phases of our experiment.
iv) \textbf{AVEC 2014} \cite{avec14}: Comprising 300 video clips from 83 individuals, both AVEC 2013 and AVEC 2014 record data in a human-computer interaction scenario. The labels, assessed using the Beck Depression Inventory-II \cite{bdi}, range from 0 to 63. Frames extracted from both datasets are cropped, aligned, and employed for training, validation, and testing based on the official split.


\subsection{Analysis of Optical Lens}

\begin{figure*}[t]
	\centering
		\centerline{\includegraphics[width=\linewidth]{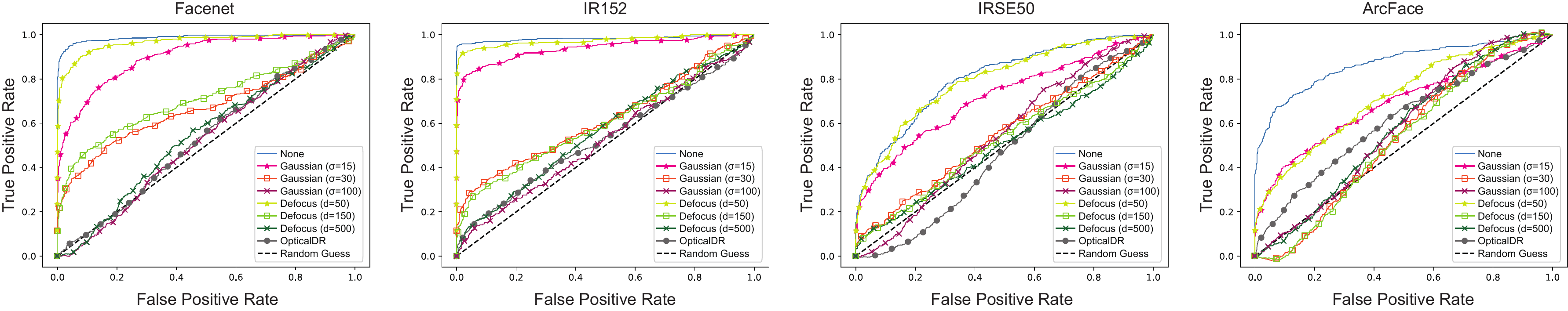}}
		\caption{ROC curves depicting the performance of facial recognition models under privacy-preserving approaches, including Gaussian blur and defocus methods, alongside our OpticalDR.}
		\label{fig_roc_ours}
\end{figure*}

\begin{figure}[t]
	\centering
		\centerline{\includegraphics[width=\linewidth]{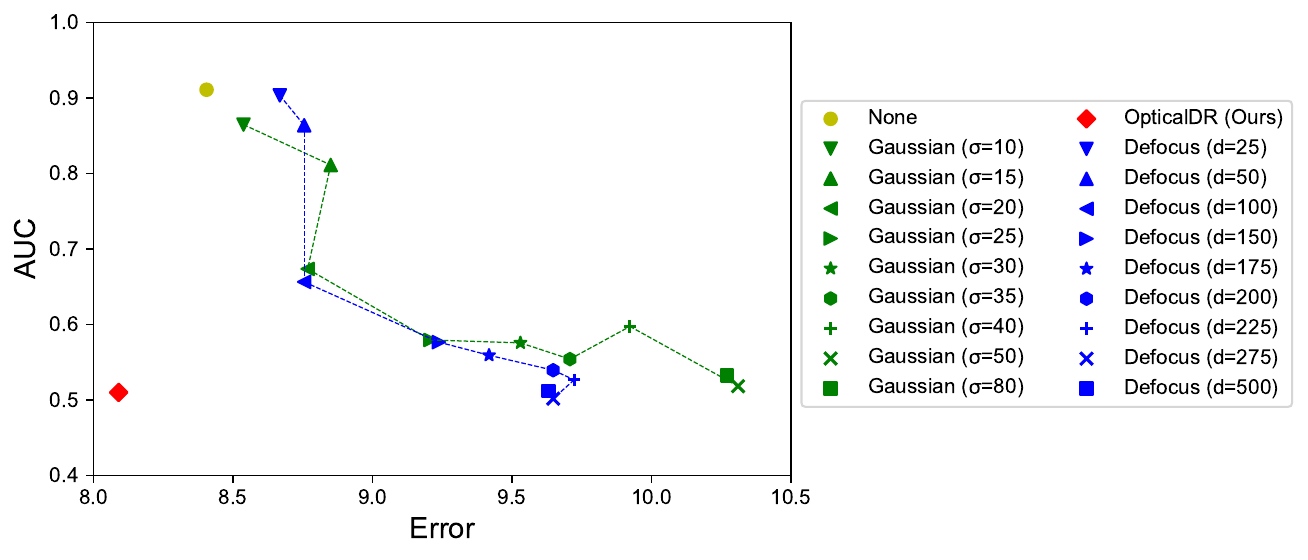}}
		\caption{Trade-off between privacy preservation and DR performance among different privacy-preserving strategies.}
		\label{fig_auc_error_part}
\end{figure}

To assess the impact of DR when employing an optimizable lens to distort images in comparison to conventional image blurring methods, we established a pipeline for images subjected to various privacy-preserving strategies. 
We conducted comparisons of the results obtained from Gaussian blurring with different $\sigma$ of the Gaussian kernel, and different defocus strengths using an approximate bokeh PSF with different bokeh ball diameters. To evaluate the privacy-preserving efficacy of each solution, we employed four widely used face recognition models: IR152 \cite{resnet}, IRSE50 \cite{8701503}, ArcFace \cite{8953658}, and Facenet \cite{7298682}, to measure the recognizability in terms of the Area Under the Curve (AUC). Furthermore, Fig.~\ref{fig_roc_ours} presents the Receiver Operating Characteristic (ROC) curves for our OpticalDR and alternative strategies. Detailed comparisons of DR and privacy-preserving performance are provided in Fig.~\ref{fig_auc_error_part}. It is observed that increasing blurring strength enhances privacy preservation (closer to random guess ROC), but this comes at the cost of a substantial decrease in DR performance. Notably, our learnable privacy-preserving approach demonstrates a significant advantage by achieving the lowest prediction error in Mean Absolute Error (MAE) and Root Mean Square Error (RMSE). This suggests that the optical lens retains more useful features even when the image is significantly distorted. 

\begin{figure*}[t]
	\centering
		\centerline{\includegraphics[width=0.9\linewidth]{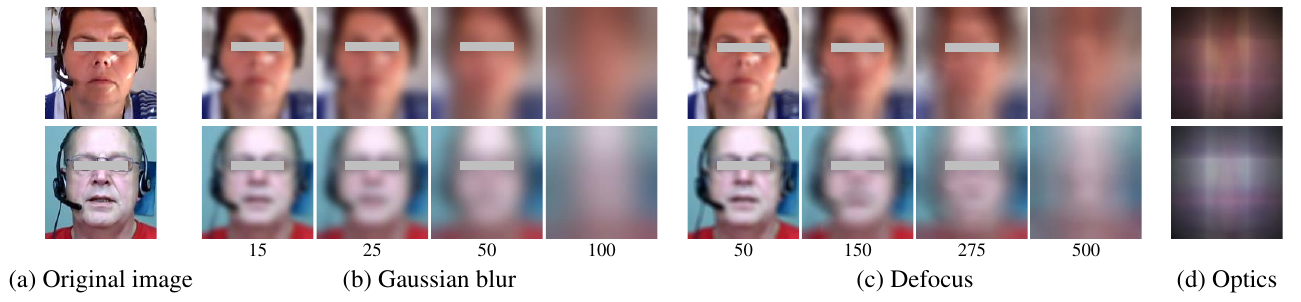}}
		\caption{The visual effects of (a) the original facial image, (b) Gaussian blur with different sigma values, (c) defocus with varying diameters of bokeh balls, and (d) the optical solution for facial privacy preservation.}
		\label{fig_blur_show}
\end{figure*}

From the colored lines in Tab.~\ref{table_exp2}, it is evident that while Gaussian blur and defocus approaches show similar performance to the optics solution in privacy preservation, the recognition error increases (yellow lines), and vice versa (blue lines). In the case of defocus with a diameter of 275 and the Gaussian blurring with $\sigma$ is 50, the privacy preserving performance is quite similar to the optics solution, but the visualization in Fig.~\ref{fig_blur_show} reveals additional identify information such as skin color and a potential gender of the subject. In the broader context of privacy, this information is also considered sensitive and should be safeguarded. In contrast, the optics solution exhibits a more robust image against various sensitive information leakage.

\begin{table*}[t]
\centering
\caption{DR performance in terms of MAE and RMSE of different privacy-preserving methods evaluated on AVEC 2013 and AVEC 2014 datasets. Additionally, privacy-preserving performance assessed on the CelebA dataset, measuring the AUC under various facial recognition models. Average error and AUC values are computed for a comprehensive comparison.}
\label{table_exp2}
\begin{tabular}{lcccc|cccc|cc}
\hline
                          & \multicolumn{2}{c}{AVEC 2013}                                & \multicolumn{2}{c|}{AVEC 2014}                               & \multicolumn{4}{c|}{AUC (CelebA)}                                                                                         &                                               &                                             \\ \cline{2-9}
\multirow{-2}{*}{Methods} & MAE                          & RMSE                          & MAE                          & RMSE                          & Facenet                      & IR152                        & IRSE50                       & ArcFace                      & \multirow{-2}{*}{$\overline{\mathrm{Error}}$} & \multirow{-2}{*}{$\overline{\mathrm{AUC}}$} \\ \hline
None                      & 7.25                         & 9.57                          & 7.27                         & 9.53                          & 0.99                         & 0.98                         & 0.80                         & 0.87                         & 8.41                                          & 0.91                                        \\
Gaussian ($\sigma$=15)    & \cellcolor[HTML]{DAE8FC}7.81 & \cellcolor[HTML]{DAE8FC}9.94  & \cellcolor[HTML]{DAE8FC}7.82 & \cellcolor[HTML]{DAE8FC}9.83  & \cellcolor[HTML]{DAE8FC}0.90 & \cellcolor[HTML]{DAE8FC}0.94 & \cellcolor[HTML]{DAE8FC}0.72 & \cellcolor[HTML]{DAE8FC}0.69 & \cellcolor[HTML]{DAE8FC}8.85                  & \cellcolor[HTML]{DAE8FC}0.81                \\
Gaussian ($\sigma$=25)    & 8.37                         & 10.06                         & 8.47                         & 9.93                          & 0.67                         & 0.61                         & 0.55                         & 0.51                         & 9.21                                          & 0.59                                        \\
Gaussian ($\sigma$=50)    & \cellcolor[HTML]{FFFFC7}9.44 & \cellcolor[HTML]{FFFFC7}11.33 & \cellcolor[HTML]{FFFFC7}9.13 & \cellcolor[HTML]{FFFFC7}11.34 & \cellcolor[HTML]{FFFFC7}0.58 & \cellcolor[HTML]{FFFFC7}0.57 & \cellcolor[HTML]{FFFFC7}0.54 & \cellcolor[HTML]{FFFFC7}0.52 & \cellcolor[HTML]{FFFFC7}10.31                 & \cellcolor[HTML]{FFFFC7}0.55                \\ \hline
Defocus ($d$=50)          & \cellcolor[HTML]{DAE8FC}7.67 & \cellcolor[HTML]{DAE8FC}9.84  & \cellcolor[HTML]{DAE8FC}7.78 & \cellcolor[HTML]{DAE8FC}9.73  & \cellcolor[HTML]{DAE8FC}0.97 & \cellcolor[HTML]{DAE8FC}0.97 & \cellcolor[HTML]{DAE8FC}0.79 & \cellcolor[HTML]{DAE8FC}0.72 & \cellcolor[HTML]{DAE8FC}8.76                  & \cellcolor[HTML]{DAE8FC}0.86                \\
Defocus ($d$=150)         & 8.13                         & 10.39                         & 8.14                         & 10.29                         & 0.71                         & 0.60                         & 0.53                         & 0.53                         & 9.24                                          & 0.59                                        \\
Defocus ($d$=275)         & \cellcolor[HTML]{FFFFC7}8.66 & \cellcolor[HTML]{FFFFC7}10.73 & \cellcolor[HTML]{FFFFC7}8.64 & \cellcolor[HTML]{FFFFC7}10.56 & \cellcolor[HTML]{FFFFC7}0.53 & \cellcolor[HTML]{FFFFC7}0.54 & \cellcolor[HTML]{FFFFC7}0.50 & \cellcolor[HTML]{FFFFC7}0.51 & \cellcolor[HTML]{FFFFC7}9.65                  & \cellcolor[HTML]{FFFFC7}0.52                \\
OpticalDR                 & 7.92                         & 8.39                          & 7.48                         & 8.57                          & 0.51                         & 0.51                         & 0.51                         & 0.51                         & 8.09                                          & 0.51                                        \\ \hline
\end{tabular}
\end{table*}

\subsection{Analysis of Privacy Robustness}

\begin{figure}[t]
	\centering
		\centerline{\includegraphics[width=0.8\linewidth]{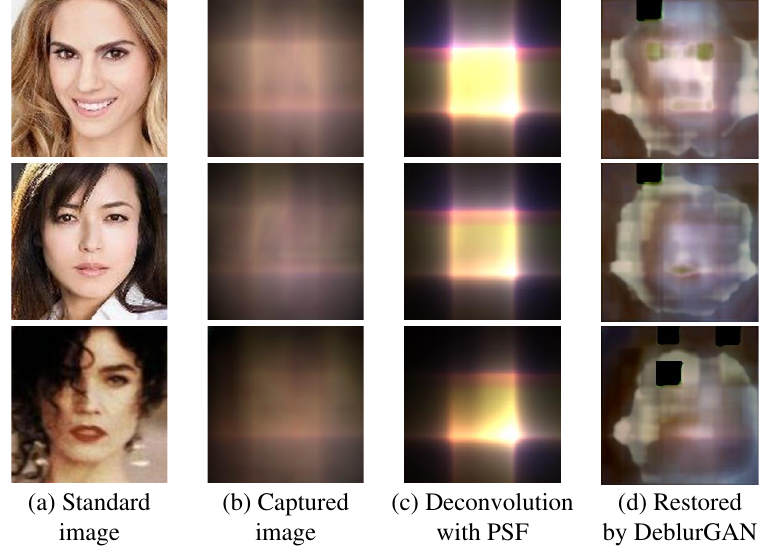}}
		\caption{A quantitative comparison of samples from CelebA is provided. The displayed images include the simulated captured images along with their corresponding deblurred results. It is noteworthy that the restored images generally exhibit poor visual recognition of human faces and appearances. }
		\label{fig_restore}
\end{figure}

We conducted a series of tests to evaluate the deconvolution method on facial images captured through the lens. These tests encompassed the following scenarios: i) \textbf{Non-blind deconvolution with known PSF:} This test simulates a scenario in which an attacker has access to the device and can obtain the PSF of the camera. We applied the non-blind Wiener deconvolution method under these conditions; ii) \textbf{Deblur GAN:} We investigated the effectiveness of blind restoration approaches using GAN-based image restoration techniques, as outlined in \cite{9008540}. This method simulates a scenario in which an attacker has access to the camera and can collect clear and blurred image pairs, but doesn't get the PSF. The attacker can employ these pairs to train a deblurring model for image recovery.

The input data for these tests consisted of blurred images and their corresponding restored images, as illustrated in Fig.~\ref{fig_restore}. It is evident that both the deconvolution and deblurring approaches exhibited limitations in restoring the images, characterized by a multitude of aberrations and significant loss of visual information.


\subsection{Analysis of Facial Features}

We explored the cumulative impact of incorporating information from identity (\textit{I}), emotions (\textit{E}), depression features (\textit{D}), and the fusion model for depression score prediction (\textit{S}). Throughout the auxiliary task, we assessed the performance of the respective tasks, and the results are presented in Tab.~\ref{table_exp4}. Regarding \emph{Step 1} (\textit{I}), our results reveal that the generated images, after training the lens, effectively erase identity information, as evidenced by an AUC of identity recognition near 0.5.

\begin{table}[t]
\centering
\caption{Performance of the proposed auxiliary tasks during the pretraining stages. \textit{I, E, D, S} represent the \emph{Step 1} to \emph{Step 4} from the previous joint optimization stages, respectively.}
\label{table_exp4}
\resizebox{\linewidth}{!}{
\begin{tabular}{lllllll}
\hline
\multirow{2}{*}{Task} & CelebA & \multicolumn{2}{l}{AVEC   2013} & \multicolumn{2}{l}{AVEC   2014} & \multirow{2}{*}{$\overline{\mathrm{Error}}$} \\ \cline{2-6}
                      & AUC    & MAE            & RMSE           & MAE            & RMSE           &                                              \\ \hline
I                     & 0.5036 & -              & -              & -              & -              & -                                            \\ \hline
D                     & 0.5352 & 10.35          & 11.28          & 10.31          & 11.25          & 10.80                                        \\
E → D                 & 0.5328 & 9.18           & 10.04          & 9.06           & 9.88           & 9.54                                         \\
I → D                 & 0.5026 & 10.15          & 11.03          & 10.27          & 11.15          & 10.65                                        \\
I → E → D             & 0.5038 & 8.54           & 9.63           & 8.87           & 9.97           & 9.25                                         \\ \hline
S                     & 0.5367 & 8.46           & 11.12          & 8.62           & 11.29          & 9.87                                         \\
I → S                 & 0.5013 & 8.24           & 10.19          & 8.60           & 10.62          & 9.41                                         \\
I → E → D → S         & 0.5016 & 7.92           & 8.39           & 7.48           & 8.57           & 8.09                                         \\ \hline
\end{tabular}
}
\end{table}

Subsequently, the emotion recognition task in \emph{Step 2} and the DR task in \emph{Step 3} involve the incorporation of emotion and depression-related information, which is utilized in the final depression score prediction. We conducted tests using four training methods at \emph{Step 3}: i) Training both the lens and the SANet from scratch (\textit{D}); ii) Training the lens from scratch while initializing the SANet with the weights obtained in \emph{Step 3} (\textit{E → D}); iii) Training the SANet from scratch while fine-tuning the lens acquired in \emph{Step 1} (\textit{I → D}); iv) Loading both the lens and SANet weights acquired in \emph{Step 2} (\textit{I → E → D}). These trained models only contain the lens and the depression SANet, as shown in \emph{Step 3} in Fig.~\ref{fig_1}. Then the models underwent evaluation to assess their performance in predicting depression scores.

Comparing cases with emotion recognition task (\textit{E → D} and \textit{I → E → D}) to those without emotion recognition pretraining information (\textit{D} and \textit{I → D}) signifies whether emotional information stored in the network is utilized in extracting depression-related features. It is observed that utilizing emotional information significantly improves the DR performance, with an approximately 10\% improvement, underscoring the importance of incorporating emotional information for extracting depression-related features.

Contrasting cases with lens pretraining (\textit{I → D} and \textit{I → E → D}) and without lens pretraining (\textit{D} and \textit{E → D}) evaluates the impact of the blend of identity features in the optical lens on DR. The incorporation of these identity features extracted by the lens can boost the recognition performance, but it is less crucial compared to the emotional features extracted from the SANet (7\% vs. 10\% improvement). This suggests that information stored within the SANet is more critical in DR than that stored in the lens, although both contribute to performance enhancement.

Furthermore, comparing the model with lens and SANet trained from scratch (\textit{D}), a noticeable performance improvement is observed when pretraining the emotion SANet and concurrently training the lens from scratch (\textit{E → D}). However, pretraining the lens while omitting information in the emotion SANet (\textit{I → D}) results in only a slight performance improvement. This underscores that the emotion model already contains advanced features suitable for recognizing images obscured by the lens, particularly beneficial for DR.

To demonstrate the necessity of the designed auxiliary tasks in OpticalDR, we conducted tests with i) training the entire model without any auxiliary task but only with $L_s$ (\textit{S}), ii) pretraining only using the privacy-preserving lens in \emph{Step 1} (\textit{I → S}), and iii) the full pretraining strategy (\textit{I → E → D → S}). Comparison of DR errors indicates the effectiveness of auxiliary tasks for improved DR performance. Moreover, AUC values in Tab.~\ref{table_exp4} show that the model maintains privacy preservation as training steps progress, evidenced by close facial recognition AUC between \textit{I → S} and \textit{I → E → D → S}. This demonstrates that training with emotional and depression information does not compromise privacy-preserving ability.



\subsection{Comparison with Facial Features Disentanglement}

Since we progressively train the OpticalDR, during which we involve the disentanglement of facial features, emotion, and depression features, we assessed the performance of TDGAN \cite{TDGAN}, which employs the disentanglement of identity and emotion features for facial expression recognition. The results are presented in Tab.~\ref{table_tdgan}. TDGAN demonstrates competitive performance, particularly for the AVEC 2014 dataset. However, TDGAN heavily depends on visibly clear facial images.

\begin{table}[t]
\centering
\scriptsize
\caption{Performance Comparison: OpticalDR vs. TDGAN. }
\label{table_tdgan}
\resizebox{1\linewidth}{!}{
\begin{tabular}{lccccc}
\hline
\multirow{2}{*}{Method} & \multirow{2}{*}{\begin{tabular}[c]{@{}c@{}}Privacy-\\ Preserving\end{tabular}} & \multicolumn{2}{c}{AVEC 2013} & \multicolumn{2}{c}{AVEC 2014} \\ \cline{3-6} 
                        &                                                                                & MAE           & RMSE          & MAE           & RMSE          \\ \hline
TDGAN      & $\times$  & 8.56  & 10.75 & 8.26  & 10.50         \\
OpticalDR  & $\surd$   & 7.53  & 8.48  & 7.89  & 8.82          \\ \hline
\end{tabular}
}
\end{table}

\subsection{Time and Memory Consumption}

Once the optical lens is optimized and deployed, the inference time and memory consumption depend solely on the deep-learning model in OpticalDR. We compared FLOPs, Params, and inference time in Tab.~\ref{table_rb1}, and our method is comparable. It's crucial to note that the handcrafted feature extraction in LQGDNet makes its inference time not directly comparable with other deep-learning methods.

\begin{table}[t]
\footnotesize
\centering
\caption{Comparison of time and memory consumption. }
\label{table_rb1}
\begin{tabular}{lccc}
\hline
Method                 & FLOPs/G & Params/M & Time/ms \\ \hline
LQGDNet \cite{9667301} & 0.23    & 2.44     & 78.63   \\
ViT \cite{vit}         & 11.29   & 58.07    & 4.15    \\
STA-DRN \cite{PAN2024121410}   & 4.84    & 33.32    & 2.63    \\
OpticalDR              & 4.36    & 67.28    & 11.64   \\ \hline
\end{tabular}
\end{table}


\subsection{Comparison with the State-of-the-Arts}

\begin{table}[t]
\centering
\caption{Comparison with previous methods for DR on the testing sets of AVEC 2013 and AVEC 2014.}
\label{table_sota}
\resizebox{\linewidth}{!}{
\begin{tabular}{lccccc}
	\hline
	\multirow{2}{*}{Method} & \multirow{2}{*}{\makecell[c]{Privacy- \\ Preserving}} & \multicolumn{2}{c}{AVEC 2013} & \multicolumn{2}{c}{AVEC 2014} \\ \cline{3-6} 
	&                          & MAE           & RMSE          & MAE           & RMSE          \\ \hline
	Baseline \cite{avec13,avec14} & $\times$     & 10.88   & 13.61   & 8.86   & 10.86      \\
	ViT \cite{vit}          & $\times$     & 8.14    & 10.67   & 8.37   & 10.59      \\
        STA-DRN (SA) \cite{PAN2024121410} & $\times$     & 7.25    & 9.57    & 7.27   & 9.53       \\	
        LQGDNet \cite{9667301}  & $\times$     & 6.38    & 8.20    & 6.08   & 7.84       \\
	 \hline
	LQGDNet + lens & $\sqrt{}\mkern-9mu{\smallsetminus}$  & 10.56  & 13.43  & 10.27  & 13.41      \\
	ViT + lens              & $\surd$      & 8.16    & 10.49    & 8.25   & 10.53      \\
	STA-DRN (SA) + lens     & $\surd$      & 8.62    & 9.53    & 8.67   & 9.68       \\
	OpticalDR                 & $\surd$      & 7.53    & 8.48    & 7.89   & 8.82         \\ \hline
\end{tabular}
}
\end{table}

We compared OpticalDR with state-of-the-art depression recognition (DR) methods on the AVEC 2013 and AVEC 2014 test sets. The comparisons include (1) \textit{ViT} \cite{vit}: A state-of-the-art universal vision model. (2) \textit{LQGDNet} \cite{9667301}: A state-of-the-art single-image facial DR approach. (3) \textit{STA-DRN (SA)} \cite{PAN2024121410}: A state-of-the-art 3D video-based DR approach utilizing the proposed spatial module. Additionally, we jointly trained these models with a lens for privacy-preserving purposes.

Analyzing Tab.~\ref{table_sota} yields key insights. 
Firstly, integrating optical components directly into state-of-the-art approaches, without employing additional learning strategies for feature enhancement, does not arise a significant improvement in performance. Furthermore, LQGDNet, the utilization of quaternion representation of features within facial images, still 
requires a visible facial image as input. Thus, we denoted this approach as $\sqrt{}\mkern-9mu{\smallsetminus}$ in Tab.~\ref{table_sota}. 
The non-differentiable nature of the quaternion feature extractor poses a challenge for joint optimization. Additionally, attempting to optically blur the global facial image can negatively impact the performance of DR.  
Conversely, when ViT is coupled with a lens, there is a slight improvement in performance, especially in terms of RMSE. This improvement indicates that optical interventions can offer additional information prior to ViT's processing, which can have a positive impact on DR.

\begin{figure}[t]
	\centering
		\centerline{\includegraphics[width=0.75\linewidth]{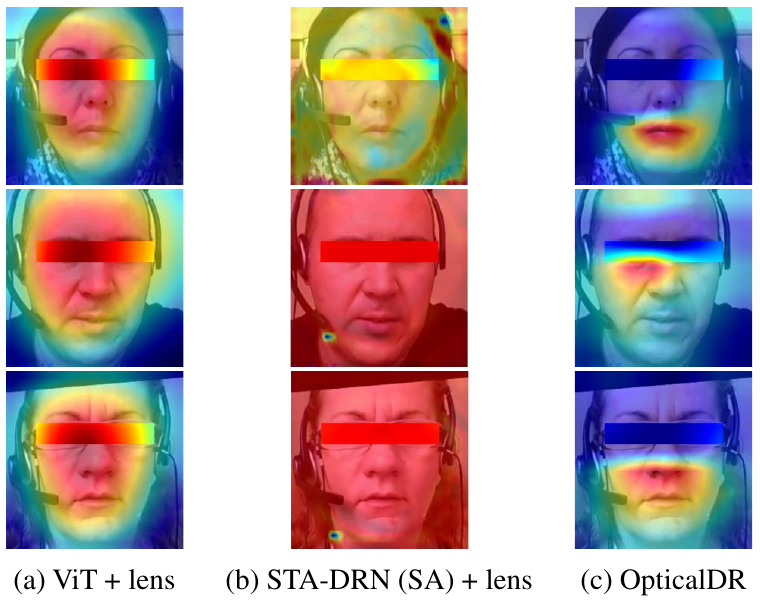}}
		\caption{CAM results for optical privacy-preserving DR approaches.}
		\label{fig_cam}
\end{figure}

Our OpticalDR achieves superior DR performance compared to \textit{ViT + lens} and \textit{STA-DRN (SA) + lens}, underscoring the effectiveness of our approach over directly joint training of optical and deep learning models. However, it falls short of outperforming LQGDNet, which leverages quaternion representation for extracting local features, still depending on visible clear facial images. This indicates that LQGDNet benefits from explicit prior information on local features like eyes and mouth. In contrast, OpticalDR, relying on the SANet, adaptively learns spatial features without the need for such explicit information, as shown in Fig.~\ref{fig_cam}.

Visualizing CAM \cite{8354201} results for OpticalDR and comparing with \textit{ViT + lens} and \textit{STA-DRN (SA) + lens} (Fig.~\ref{fig_cam}), OpticalDR actives on specific regions with clear borders, potentially containing features for DR. Contrastingly, the simple combination of optical and vision deep models results in nonsensical or unchanging activation on the facial area, underscoring the necessity and efficiency of OpticalDR's joint optimization strategy.

Overall, OpticalDR has proven to be the most effective solution in terms of privacy preservation and DR. It successfully protects facial privacy while achieving low MAE and RMSE, demonstrating its outstanding performance in DR tasks.


\section{Conclusion}

We present OpticalDR, an innovative privacy-preserving method for DR. Unlike previous image-based approaches, OpticalDR prioritizes capturing privacy-preserving images using an optimized lens to preserve depression-related features while concealing identity information. Crucially, OpticalDR operates without relying on visually identifiable facial images during system inference, ensuring robust privacy protection. Experimental results showcase OpticalDR's state-of-the-art privacy-preserving performance and competitive DR recognition on established benchmarks. In the future, our efforts will involve the physical deployment of the entire system and the integration of additional modalities, including temporal information, EEG data, and consultation transcripts, all within the framework designed for privacy preservation.

{\small
\bibliographystyle{ieee_fullname}
\bibliography{egbib}

\begin{thebibliography}{10}\itemsep=-1pt

\bibitem{2.4}
Sharifa Alghowinem, Roland Goecke, Michael Wagner, Julien Epps, Matthew Hyett, Gordon Parker, and Michael Breakspear.
\newblock Multimodal depression detection: Fusion analysis of paralinguistic, head pose and eye gaze behaviors.
\newblock {\em IEEE Transactions on Affective Computing}, 9(4):478--490, 2018.

\bibitem{bdi}
Aaron~T. Beck, Robert~A. Steer, Roberta Ball, and William~F. Ranieri.
\newblock Comparison of beck depression inventories-{IA} and-{II} in psychiatric outpatients.
\newblock {\em Journal of Personality Assessment}, 67(3):588--597, 1996.

\bibitem{9186792}
Wheidima Carneiro~de Melo, Eric Granger, and Abdenour Hadid.
\newblock A deep multiscale spatiotemporal network for assessing depression from facial dynamics.
\newblock {\em IEEE Transactions on Affective Computing}, pages 1--1, 2020.

\bibitem{Chang2018}
Julie Chang, Vincent Sitzmann, Xiong Dun, Wolfgang Heidrich, and Gordon Wetzstein.
\newblock Hybrid optical-electronic convolutional neural networks with optimized diffractive optics for image classification.
\newblock {\em Scientific Reports}, 8(1):12324, Aug 2018.

\bibitem{9010976}
Julie Chang and Gordon Wetzstein.
\newblock Deep optics for monocular depth estimation and 3{D} object detection.
\newblock In {\em IEEE/CVF International Conference on Computer Vision (ICCV)}, pages 10192--10201, 2019.

\bibitem{8354201}
Aditya Chattopadhay, Anirban Sarkar, Prantik Howlader, and Vineeth~N Balasubramanian.
\newblock {Grad-CAM++}: Generalized gradient-based visual explanations for deep convolutional networks.
\newblock In {\em IEEE Winter Conference on Applications of Computer Vision (WACV)}, pages 839--847, 2018.

\bibitem{8953658}
Jiankang Deng, Jia Guo, Niannan Xue, and Stefanos Zafeiriou.
\newblock Arc{F}ace: Additive angular margin loss for deep face recognition.
\newblock In {\em IEEE/CVF Conference on Computer Vision and Pattern Recognition (CVPR)}, pages 4685--4694, 2019.

\bibitem{DONG2021279}
Yizhuo Dong and Xinyu Yang.
\newblock A hierarchical depression detection model based on vocal and emotional cues.
\newblock {\em Neurocomputing}, 441:279--290, 2021.

\bibitem{vit}
Alexey {Dosovitskiy}, Lucas {Beyer}, Alexander {Kolesnikov}, Dirk {Weissenborn}, Xiaohua {Zhai}, Thomas {Unterthiner}, Mostafa {Dehghani}, Matthias {Minderer}, Georg {Heigold}, Sylvain {Gelly}, Jakob {Uszkoreit}, and Neil {Houlsby}.
\newblock An image is worth 16x16 words: Transformers for image recognition at scale.
\newblock {\em arXiv e-prints}, page arXiv:2010.11929, Oct. 2020.

\bibitem{8756584}
Zhengyin Du, Weixin Li, Di Huang, and Yunhong Wang.
\newblock Encoding visual behaviors with attentive temporal convolution for depression prediction.
\newblock In {\em IEEE International Conference on Automatic Face Gesture Recognition (FG)}, pages 1--7, 2019.

\bibitem{2018arXiv181108592H}
Albert {Haque}, Michelle {Guo}, Adam~S {Miner}, and Li {Fei-Fei}.
\newblock Measuring depression symptom severity from spoken language and {3D} facial expressions.
\newblock {\em arXiv e-prints}, page arXiv:1811.08592, Nov. 2018.

\bibitem{resnet}
Kaiming He, Xiangyu Zhang, Shaoqing Ren, and Jian Sun.
\newblock Deep residual learning for image recognition.
\newblock In {\em IEEE Conference on Computer Vision and Pattern Recognition (CVPR)}, pages 770--778, 2016.

\bibitem{8501575}
Lang He, Dongmei Jiang, and Hichem Sahli.
\newblock Automatic depression analysis using dynamic facial appearance descriptor and dirichlet process fisher encoding.
\newblock {\em IEEE Transactions on Multimedia}, 21(6):1476--1486, 2019.

\bibitem{10096627}
Chris Henry, M.~Salman Asif, and Zhu Li.
\newblock Privacy preserving face recognition with lensless camera.
\newblock In {\em IEEE International Conference on Acoustics, Speech and Signal Processing (ICASSP)}, pages 1--5, 2023.

\bibitem{optics02}
Carlos Hinojosa, Miguel Marquez, Henry Arguello, Ehsan Adeli, Li Fei-Fei, and Juan~Carlos Niebles.
\newblock {PrivHAR}: Recognizing human actions from privacy-preserving lens.
\newblock In {\em European Conference on Computer Vision (ECCV)}, page 314–332, 2022.

\bibitem{9710270}
Carlos Hinojosa, Juan~Carlos Niebles, and Henry Arguello.
\newblock Learning privacy-preserving optics for human pose estimation.
\newblock In {\em IEEE/CVF International Conference on Computer Vision (ICCV)}, pages 2553--2562, 2021.

\bibitem{8701503}
Jie Hu, Li Shen, Samuel Albanie, Gang Sun, and Enhua Wu.
\newblock Squeeze-and-excitation networks.
\newblock {\em IEEE Transactions on Pattern Analysis and Machine Intelligence}, 42(8):2011--2023, 2020.

\bibitem{9466261}
Hayato Ikoma, Cindy~M. Nguyen, Christopher~A. Metzler, Yifan Peng, and Gordon Wetzstein.
\newblock Depth from defocus with learned optics for imaging and occlusion-aware depth estimation.
\newblock In {\em IEEE International Conference on Computational Photography (ICCP)}, pages 1--12, 2021.

\bibitem{optics04}
Yasunori Ishii, Satoshi Sato, and Takayoshi Yamashita.
\newblock Privacy-aware face recognition with lensless multi-pinhole camera.
\newblock In {\em European Conference on Computer Vision (ECCV)}, page 476–493, 2020.

\bibitem{9008540}
Orest Kupyn, Tetiana Martyniuk, Junru Wu, and Zhangyang Wang.
\newblock {DeblurGAN-v2}: Deblurring (orders-of-magnitude) faster and better.
\newblock In {\em IEEE/CVF International Conference on Computer Vision (ICCV)}, pages 8877--8886, 2019.

\bibitem{9054420}
Zhi Li, Haoliang Li, Kwok-Yan Lam, and Alex~Chichung Kot.
\newblock Unseen face presentation attack detection with hypersphere loss.
\newblock In {\em IEEE International Conference on Acoustics, Speech and Signal Processing (ICASSP)}, pages 2852--2856, 2020.

\bibitem{celeba}
Ziwei Liu, Ping Luo, Xiaogang Wang, and Xiaoou Tang.
\newblock Deep learning face attributes in the wild.
\newblock In {\em IEEE International Conference on Computer Vision (ICCV)}, pages 3730--3738, 2015.

\bibitem{5543262}
Patrick Lucey, Jeffrey~F. Cohn, Takeo Kanade, Jason Saragih, Zara Ambadar, and Iain Matthews.
\newblock The extended cohn-kanade dataset ({CK+}): A complete dataset for action unit and emotion-specified expression.
\newblock In {\em IEEE Computer Society Conference on Computer Vision and Pattern Recognition - Workshops (CVPRW)}, pages 94--101, 2010.

\bibitem{optics05}
Robert~J. Noll.
\newblock Zernike polynomials and atmospheric turbulence$\ast$.
\newblock {\em Journal of the Optical Society of America}, 66(3):207--211, 1976.

\bibitem{PAN2024121410}
Yuchen Pan, Yuanyuan Shang, Tie Liu, Zhuhong Shao, Guodong Guo, Hui Ding, and Qiang Hu.
\newblock Spatial–temporal attention network for depression recognition from facial videos.
\newblock {\em Expert Systems with Applications}, 237:121410, 2024.

\bibitem{10185131}
Yuchen Pan, Yuanyuan Shang, Zhuhong Shao, Tie Liu, Guodong Guo, and Hui Ding.
\newblock Integrating deep facial priors into landmarks for privacy preserving multimodal depression recognition.
\newblock {\em IEEE Transactions on Affective Computing}, pages 1--8, 2023.

\bibitem{E17V2}
Swati Rathi, Baljeet Kaur, and R.~K. Agrawal.
\newblock Enhanced depression detection from facial cues using univariate feature selection techniques.
\newblock In {\em International Conference on Pattern Recognition and Machine Intelligence (PReMI)}, pages 22--29, 2019.

\bibitem{avec17}
Fabien Ringeval, Bj\"{o}rn Schuller, Michel Valstar, Jonathan Gratch, Roddy Cowie, Stefan Scherer, Sharon Mozgai, Nicholas Cummins, Maximilian Schmitt, and Maja Pantic.
\newblock {AVEC} 2017: Real-life depression, and affect recognition workshop and challenge.
\newblock In {\em Annual Workshop on Audio/Visual Emotion Challenge (AVEC)}, page 3–9, 2017.

\bibitem{7298682}
Florian Schroff, Dmitry Kalenichenko, and James Philbin.
\newblock Face{N}et: A unified embedding for face recognition and clustering.
\newblock In {\em IEEE Conference on Computer Vision and Pattern Recognition (CVPR)}, pages 815--823, 2015.

\bibitem{9667301}
Yuanyuan Shang, Yuchen Pan, Xiao Jiang, Zhhong Shao, Guodong Guo, Tie Liu, and Hui Ding.
\newblock {LQGDNet}: A local quaternion and global deep network for facial depression recognition.
\newblock {\em IEEE Transactions on Affective Computing}, pages 1--1, 2021.

\bibitem{10097883}
Jian Shen, Yanan Zhang, Huajian Liang, Zeguang Zhao, Kexin Zhu, Kun Qian, Qunxi Dong, Xiaowei Zhang, and Bin Hu.
\newblock Depression recognition from {EEG} signals using an adaptive channel fusion method via improved focal loss.
\newblock {\em IEEE Journal of Biomedical and Health Informatics}, 27(7):3234--3245, 2023.

\bibitem{optics01}
Vincent Sitzmann, Steven Diamond, Yifan Peng, Xiong Dun, Stephen Boyd, Wolfgang Heidrich, Felix Heide, and Gordon Wetzstein.
\newblock End-to-end optimization of optics and image processing for achromatic extended depth of field and super-resolution imaging.
\newblock {\em ACM Transactions on Graphics}, 37(4), jul 2018.

\bibitem{8976305}
Siyang Song, Shashank Jaiswal, Linlin Shen, and Michel Valstar.
\newblock Spectral representation of behaviour primitives for depression analysis.
\newblock {\em IEEE Transactions on Affective Computing}, pages 1--1, 2020.

\bibitem{8373825}
Siyang Song, Linlin Shen, and Michel Valstar.
\newblock Human behaviour-based automatic depression analysis using hand-crafted statistics and deep learned spectral features.
\newblock In {\em IEEE International Conference on Automatic Face Gesture Recognition (FG)}, pages 158--165, 2018.

\bibitem{E17V1}
Bo Sun, Yinghui Zhang, Jun He, Lejun Yu, Qihua Xu, Dongliang Li, and Zhaoying Wang.
\newblock A random forest regression method with selected-text feature for depression assessment.
\newblock In {\em Annual Workshop on Audio/Visual Emotion Challenge (AVEC)}, page 61–68, 2017.

\bibitem{unalign_trans}
Yao-Hung~Hubert Tsai, Shaojie Bai, Paul~Pu Liang, J.~Zico Kolter, Louis-Philippe Morency, and Ruslan Salakhutdinov.
\newblock Multimodal transformer for unaligned multimodal language sequences.
\newblock In {\em Annual Meeting of the Association for Computational Linguistics (ACL)}, pages 6558--6569, 2019.

\bibitem{8976084}
Md~Azher Uddin, Joolekha~Bibi Joolee, and Young-Koo Lee.
\newblock Depression level prediction using deep spatiotemporal features and multilayer bi-ltsm.
\newblock {\em IEEE Transactions on Affective Computing}, pages 1--1, 2020.

\bibitem{histloss}
Evgeniya Ustinova and Victor Lempitsky.
\newblock Learning deep embeddings with histogram loss.
\newblock In {\em International Conference on Neural Information Processing Systems (NIPS)}, page 4177–4185, 2016.

\bibitem{avec14}
Michel Valstar, Bj\"{o}rn Schuller, Kirsty Smith, Timur Almaev, Florian Eyben, Jarek Krajewski, Roddy Cowie, and Maja Pantic.
\newblock {AVEC} 2014: {3D} dimensional affect and depression recognition challenge.
\newblock In {\em International Workshop on Audio/Visual Emotion Challenge (AVEC)}, page 3–10, 2014.

\bibitem{avec13}
Michel Valstar, Bj\"{o}rn Schuller, Kirsty Smith, Florian Eyben, Bihan Jiang, Sanjay Bilakhia, Sebastian Schnieder, Roddy Cowie, and Maja Pantic.
\newblock {AVEC} 2013: The continuous audio/visual emotion and depression recognition challenge.
\newblock In {\em ACM International Workshop on Audio/Visual Emotion Challenge (AVEC)}, page 3–10, 2013.

\bibitem{TDGAN}
Siyue Xie, Haifeng Hu, and Yizhen Chen.
\newblock Facial expression recognition with two-branch disentangled generative adversarial network.
\newblock {\em IEEE Transactions on Circuits and Systems for Video Technology}, 31(6):2359--2371, 2021.

\bibitem{4.13}
Ziping Zhao, Zhongtian Bao, Zixing Zhang, Nicholas Cummins, Haishuai Wang, and Björn Schuller.
\newblock Hierarchical attention transfer networks for depression assessment from speech.
\newblock In {\em IEEE International Conference on Acoustics, Speech and Signal Processing (ICASSP)}, pages 7159--7163, 2020.

\bibitem{8344107}
Xiuzhuang Zhou, Kai Jin, Yuanyuan Shang, and Guodong Guo.
\newblock Visually interpretable representation learning for depression recognition from facial images.
\newblock {\em IEEE Transactions on Affective Computing}, 11(3):542--552, 2020.

\bibitem{9187982}
Xiuzhuang Zhou, Zeqiang Wei, Min Xu, Shan Qu, and Guodong Guo.
\newblock Facial depression recognition by deep joint label distribution and metric learning.
\newblock {\em IEEE Transactions on Affective Computing}, pages 1--1, 2020.

\end{thebibliography}
}

\end{document}


\title{OpticalDR: A Deep Optical Imaging Model for Privacy-Protective \\Depression Recognition - Supplemental Material}  

\maketitle
\thispagestyle{empty}
\appendix


\section{Training Setup}

\subsection{Lens Parameters Setup} 

In our simulation, the wavefront is modeled with a pixel size of 3.69 $\mu$m and a resolution of 1024 $\times$ 1024. We use the first 15 terms in Noll notation to represent the Zernike coefficients, which model the surface profile. The refractive indices are defined as 1.488, 1.493, and 1.499 to simulate the lens's refraction of red, green, and blue light at wavelengths of 640 nm, 550 nm, and 460 nm, respectively. The depth is randomly sampled from a range from 0.33 to 2 meters to account for potential variations in distance between individuals and the camera in real-world scenarios. The distance between the lens and sensors $z$ is set at 35.5 mm.

\subsection{Initialization, Training and Fine-Tuning Setup} 

The training of OpticalDR consists of four steps. The training details of each steps are list as follows:

\textbf{Step 1.} We utilize the CelebA dataset \cite{celeba} to train the lens and SANet, allowing the lens to generate privacy-preserving images with $L_{i}$. The lens is initialized with the fourth Zernike coefficient set to -51 and is trained using the Adadelta optimizer with a learning rate of 1. Concurrently, SANet is trained with the Adam optimizer employing a learning rate of 0.01. We save all parameters when the validation loss reaches a minimum. The SANet structure used in this stage is a compact version of ResNet, specifically ResNet10, with bottleneck blocks replaced by SA modules.

\textbf{Step 2.} The CK+ dataset \cite{5543262} is employed to train the lens and a SANet for acquiring emotional information. We use pretrained parameters of the lens and SANet from \emph{Step 1} as the initial lens and the initial emotion recognition model. The learning rate for the lens is set to 0.01, while for SANet, it is set to 0.0001 during this stage. Parameters are saved when the validation loss $L_e$ reaches a minimum.

\textbf{Step 3.} The AVEC 2014 dataset \cite{avec14} is utilized for acquire depression-related features optimizing $L_d$. Videos from AVEC 2014 are extracted frame-by-frame and employed for training purposes. During the validation process, videos are sampled with a frame interval of 10, and the average output from all frames within a single video sample is used as the result for that sample. Human face alignment is performed using the Dlib toolkit. During alignment, we ensure that the centers between the eyes are aligned, and the vertical distance between the eyes and the mouth is set to be 1/3 of the image height. In the training phase, we use the pretrained lens and SANet parameters obtained from \emph{Step 2} as the initial model weights containing emotion information. The learning rate for the lens is set to 0.01, and for SANet, it is set to 0.0001 in this stage. All parameters are saved when the validation loss $L_d$ reaches a minimum.

\textbf{Step 4.} For fine-tuning with the final fusion layer, we utilize the AVEC 2014 dataset and employ the depression self-evaluation score of each sample provided by the dataset as the label. In this step, we employ the fusion model with 2 MulT \cite{unalign_trans} layers, each having 4 attention heads. The lens uses the weight from \emph{Step 3}, and the emotion SANet and depression SANet use weights from \emph{Step 2} and \emph{Step 3}, respectively. Then the parameters of lens and SANets are frozen during training in this step. Training of the fusion layer is performed using Adam as the optimizer with a learning rate of 0.0001 for optimizing $L_s$. During validation and testing, we assess performance using MAE and RMSE as the evaluation criteria on AVEC 2013 and AVEC 2014, allowing for comparisons with other approaches. The calculation method for validation and testing involves averaging the output from frames sampled at intervals of 10.

\section{Visualization of Privacy Preserving Images}

\begin{figure*}[t]
\centering
		\centerline{\includegraphics[width=\linewidth]{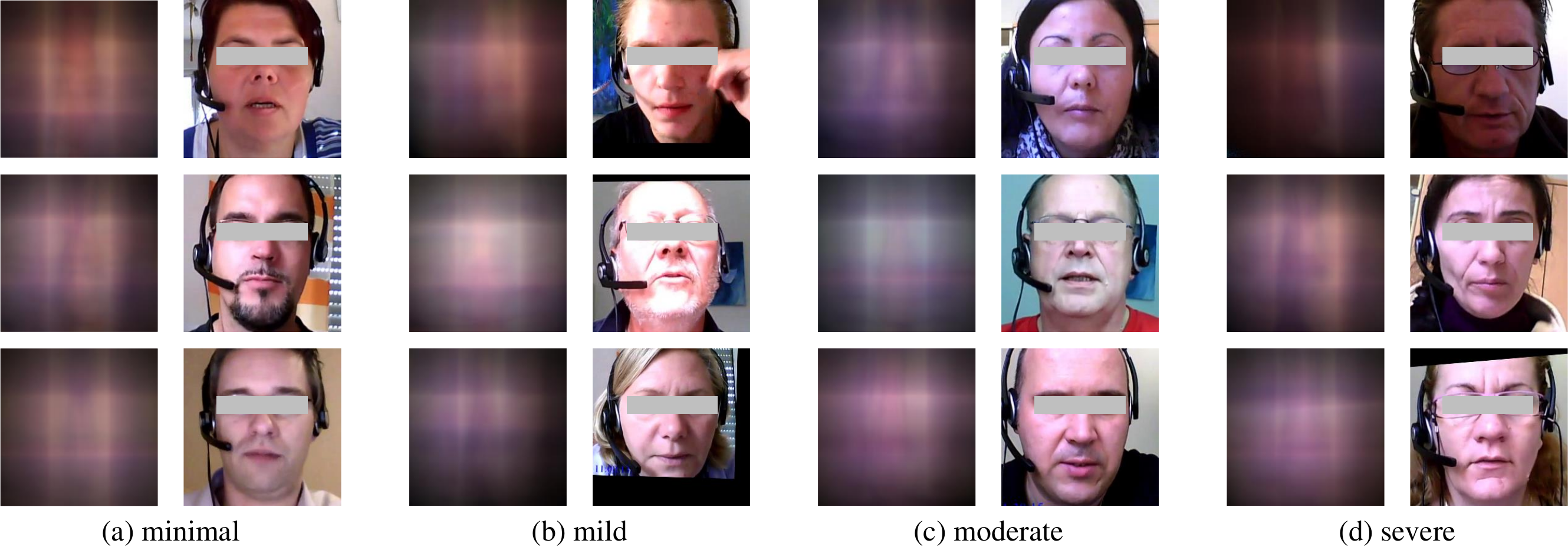}}
		\caption{The visualization of privacy-preserving images with different depression levels.}
		\label{fig_avec_cls}
\end{figure*}

In this section, we present images generated by the lens within OpticalDR, showcasing samples with diverse levels of depression, as illustrated in Fig. \ref{fig_avec_cls}. While the privacy-preserving images obscure identity details, discernible patterns persist. These patterns are likely intrinsic information utilized by the deep learning model for DR.

\section{Visualization of Lens}

\begin{figure*}[t]
\centering
		\centerline{\includegraphics[width=\linewidth]{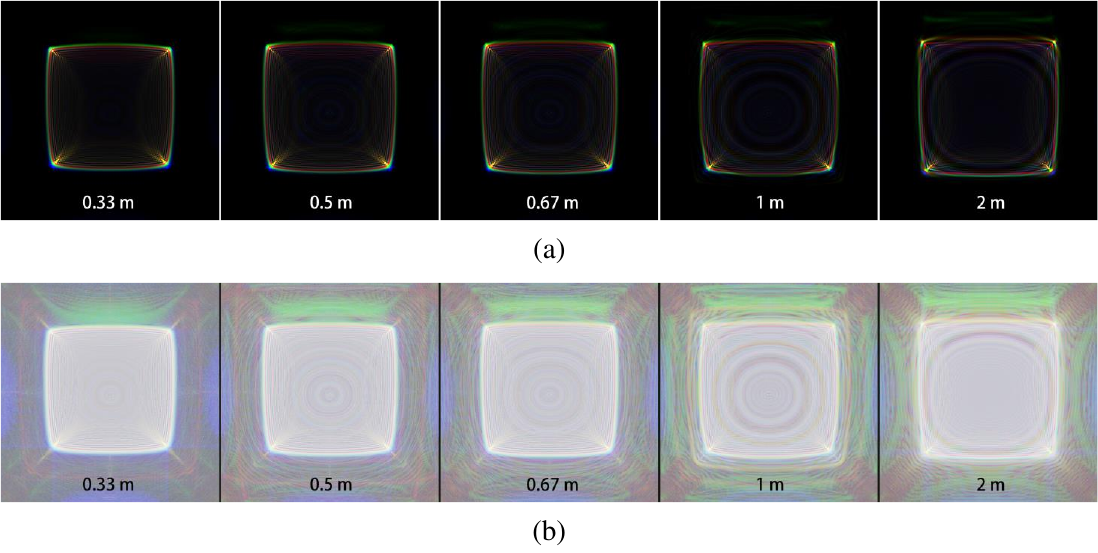}}
		\caption{Visualization of (a) PSFs of the optimized lens in OpticalDR under various focal distances and (b) Log-transformed PSFs for enhanced clarity.}
		\label{fig_psf}
\end{figure*}

We visualize the PSFs of the optimized lens in OpticalDR, as shown in Fig. \ref{fig_psf}.

{\small
\bibliographystyle{ieee_fullname}
\bibliography{egbib}
}